\documentclass[letterpaper]{article} 

\usepackage{aaai2026}  
\usepackage{times}  
\usepackage{helvet}  
\usepackage{courier}  
\usepackage[hyphens]{url}  
\usepackage{graphicx} 
\usepackage{subcaption}
\usepackage{natbib}  
\usepackage{caption} 
\usepackage{url}
\usepackage{xcolor}
\usepackage[dvipsnames]{xcolor}
\usepackage[table]{xcolor}
\usepackage{algorithm}
\usepackage{algorithmic}
\usepackage{booktabs}
\usepackage{newfloat}
\usepackage{listings}
\DeclareCaptionStyle{ruled}{labelfont=normalfont,labelsep=colon,strut=off} 
\floatstyle{ruled}
\newfloat{listing}{tb}{lst}{}
\floatname{listing}{Listing}
\title{DiffBench Meets DiffAgent: End‑to‑End LLM‑Driven Diffusion Acceleration Code Generation}

\author{
    Jiajun jiao\textsuperscript{\rm 1, 2}\equalcontrib, Haowei Zhu\textsuperscript{\rm 1,3}\equalcontrib, Puyuan Yang\textsuperscript{\rm 1}, Jianghui Wang\textsuperscript{\rm 1}, Ji Liu\textsuperscript{\rm 1}, Ziqiong Liu\textsuperscript{\rm 1}, Dong Li\textsuperscript{\rm 1}, Yuejian Fang\textsuperscript{\rm 2}\thanks{Corresponding author.}, Junhai Yong\textsuperscript{\rm 3}, Bin Wang\textsuperscript{\rm 3}\footnotemark[2], Emad Barsoum\textsuperscript{\rm 1}
}
\affiliations{
    \textsuperscript{\rm 1} Advanced Micro Devices, Inc.\\

    \textsuperscript{\rm 2} Peking University \\
    \textsuperscript{\rm 3} Tsinghua University

    \{jiajjiao, haoweiz, puyuyang, jianghui.wang, liuji, ziqioliu, d.li, ebarsoum\}@amd.com, fangyj@ss.pku.edu.cn, yongjh@tsinghua.edu.cn, wangbins@tsinghua.edu.cn
}

\usepackage{bibentry}
\usepackage{amsmath}

\begin{document}

\maketitle

\begin{abstract}
Diffusion models have achieved remarkable success in image and video generation. However, their inherently multiple step inference process imposes substantial computational overhead, hindering real-world deployment. Accelerating diffusion models is therefore essential, yet determining how to combine multiple model acceleration techniques remains a significant challenge. To address this issue, we introduce a framework driven by large language models (LLMs) for automated acceleration code generation and evaluation. First, we present DiffBench, a comprehensive benchmark that implements a three stage automated evaluation pipeline across diverse diffusion architectures, optimization combinations and deployment scenarios. Second, we propose DiffAgent, an agent that generates optimal acceleration strategies and codes for arbitrary diffusion models. DiffAgent employs a closed-loop workflow in which a planning component and a debugging component iteratively refine the output of a code generation component, while a genetic algorithm extracts performance feedback from the execution environment to guide subsequent code refinements. We provide a detailed explanation of the DiffBench construction and the design principles underlying DiffAgent. Extensive experiments show that DiffBench offers a thorough evaluation of generated codes and that DiffAgent significantly outperforms existing LLMs in producing effective diffusion acceleration strategies.

\end{abstract}
\section{Introduction}

Diffusion models \cite{podell2023sdxl, esser2024scaling, batifol2025flux} have rapidly become the preferred approach for high fidelity generative tasks in computer vision \cite{zhu2024distribution, wan2025wan, zhu2025recon}. Their iterative sampling procedures, however, introduce considerable latency and computational overhead. Although a variety of acceleration strategies have been proposed to reduce inference cost \cite{lu2022dpm, ma2024deepcache, liu2025fasterdiffusiontemporalattention, bolya2023token}, these methods require expert intervention and bespoke engineering for each model architecture and deployment scenario. As diffusion architectures diversify to include U‑Net variants \cite{ronneberger2015unetconvolutionalnetworksbiomedical} and transformer based models \cite{peebles2023scalable, dosovitskiy2021an} and deployment contexts become more varied, there is a pressing need for automated tools capable of navigating this complex design space and producing correct, efficient implementation code without manual effort.

Recent advances in large language models (LLMs) have demonstrated their potential for end to end code generation and optimization \cite{zhang2024codeagent}. Domain-aware LLM agents applied to specialized tasks, such as GPU kernel optimization benchmarks \cite{kernelbench2025, li2025tritonbench}, have achieved notable performance gains over naive implementations. Existing research, however, has not confronted the unique challenges of generating diffusion inference code, in particular the integration of diffusion architectures with tailored acceleration techniques while satisfying strict performance and accuracy constraints in target deployment scenarios. Developers currently rely on extensive documentation libraries and hardware specific evaluations to ensure that implementations meet both precision and efficiency requirements. Crafting such solutions demands deep expertise in diffusion modeling, acceleration methods and parameter tuning, capabilities that are beyond the reach of current LLMs.

Diffusion acceleration code generation differs from general code synthesis because it requires substantial domain knowledge and environmental feedback to evaluate and refine implementations. To address these needs, we first design an environment that emulates a human developer’s workflow and supports comprehensive evaluation of LLM outputs. This environment automates the selection and tuning of acceleration techniques across a wide range of diffusion algorithms and hardware platforms and assesses both functional correctness and performance. We introduce DiffBench, a benchmark consisting of 604 prompts and reference implementations drawn from real world deployment scenarios and covering diverse model architectures, control conditions and acceleration strategies. DiffBench defines a three stage evaluation protocol that enables rigorous assessment of LLM performance.

In addaition, building on human programming practice, in which developers iteratively reflect on feedback to adjust their code, we propose DiffAgent, an LLM based agent framework for diffusion acceleration code generation. DiffAgent integrates planning \cite{wei2023chainofthoughtpromptingelicitsreasoning, jiang2024selfplanningcodegenerationlarge}, coding  and debugging \cite{Li_2022, chen2023teachinglargelanguagemodels} stages and employs a genetic algorithm based selector to generate code and guide successive refinements. We validate DiffAgent on DiffBench using several state-of-the-art LLMs including  GPT-4.1, Claude Sonnet 4, Gemeni 2.5 Flash \cite{comanici2025gemini} and o3-mini. Compared to direct code generation by existing models, DiffAgent yields improvements of 54.30\% to 81.59\% across all benchmarks. A detailed case analysis shows that our framework significantly reduces common error modes in LLM generated code. 
Our main contributions are summarized as follows:
\begin{itemize}
  \item We formalize the problem of diffusion acceleration code generation, providing a structured framework to evaluate the ability of large language models to autonomously produce optimized diffusion inference code for real‑world deployment scenarios.
  \item We introduce DiffBench, a benchmark for diffusion acceleration code generation that comprises high‑quality code repositories and spans a wide variety of real‑world deployment scenarios and acceleration strategies.
  \item We propose DiffAgent, an agentic framework built on large language models that orchestrates planning, coding, debugging and evaluation stages and employs a genetic algorithm to tune implementation parameters in order to meet prescribed accuracy and efficiency targets.
  \item  We conduct comprehensive experiments on four large language models to demonstrate the versatility and effectiveness of DiffAgent in generating diffusion acceleration code and to quantify its performance improvements under practical constraints.
\end{itemize}

\section{Related Work}

\subsection{Diffusion Model Acceleration}
Recent progress in diffusion model acceleration targets both algorithmic and system-level optimizations to enable real-time high-quality generation \cite{rombach2022high}. On the algorithmic front, learnable gating mechanisms \cite{tgate} evaluate a scalar score at each denoising step and conditionally skip redundant updates, reducing network calls by nearly forty percent with minimal impact on visual fidelity. Complementing this, feature-level caching frameworks \cite{ma2024deepcache, dipgo, qin2025accelerating} store and reuse early UNet activations across adjacent timesteps, exploiting their slow variation to avoid redundant convolutions and achieve up to fifty percent speedups alongside lower memory peaks. From the transformer world, token merging techniques \cite{bolya2023token} iteratively fuse semantically similar feature tokens within self-attention blocks, halving spatial resolution while preserving content and doubling throughput on high-resolution outputs \cite{ryali2023hierahierarchicalvisiontransformer} with only slight PSNR degradation. Finally, just-in-time compilation APIs (e.g., torch compile) fuse elementwise operations, reorder memory accesses, and eliminate Python overhead in the denoising network, delivering twenty to thirty-five percent faster sampling across a wide range of diffusion checkpoints. By uniting adaptive skipping, activation reuse, token merging, and kernel-level compilation, these approaches jointly push the efficiency frontier for diffusion-based generative modeling.

\subsection{Benchmarks and Evaluation}
Evaluation methodologies for generative models have evolved to emphasize task-agnostic rigor and multi-dimensional assessment \cite{borji2021prosconsganevaluation, maekawa2025holistic}. Low-level code-generation benchmarks \cite{kernelbench2025, li2025tritonbench, pan2025codevbench} measure a language model's capacity to synthesize high-performance GPU kernels, providing insight into fine-grained computational reasoning. In multimodal generation, learned alignment metrics \cite{radford2021clip, liu2024alignbenchbenchmarkingchinesealignment} quantify the correspondence between text prompts and synthesized images, while domain-specific criteria, such as transparency-based metrics for surface-defect detection \cite{zhang2023transfusion}, offer targeted evaluations of visual fidelity. Three-dimensional reconstruction tasks further broaden the scope: event-based data augmentation protocols \cite{xu2023eventzoom} stress neuromorphic vision pipelines with progressive stimuli, and point-level anomaly detection benchmarks assess performance on irregular 3D structures. Collectively, these benchmarks ensure that advances in speed and model architecture are balanced by comprehensive, multi-axis validation.

\subsection{Agent-based Systems}

The field of agent-based systems is undergoing a significant transformation, largely driven by the integration of large language models (LLMs). This evolution is prominent in multi-agent coordination and LLM-driven decision-making \cite{xue2024multi, li2025agentoriented, schmidgall2025agentlaboratoryusingllm, dong2024selfcollaborationcodegenerationchatgpt}. Within multi-agent systems, novel approaches such as structural-information objectives are enabling agents to achieve sophisticated role discovery in complex environments.

A promising frontier is the rise of specialized coding agents that exploit the high-level reasoning capabilities of large language models to autonomously plan, generate, and iteratively refine software solutions \cite{zhang2025cumulativereasoninglargelanguage, islam2024mapcodermultiagentcodegeneration, huang2024agentcodermultiagentbasedcodegeneration, guo2024deepseekcoderlargelanguagemodel}. These agents collaborate, decompose tasks, write tests, and optimize performance, consistently outperforming conventional baselines in both quality and speed. Reinforcement learning can play a supporting role (e.g., aligning low-level decision policies with latent textual rewards \cite{kim2023lare}), but the principal innovation lies in coupling LLM-driven natural-language reasoning with domain-specific tool use \cite{wölflein2025llmagentsmakingagent} and long-term memory \cite{xu2025amemagenticmemoryllm, wang2025agent} to enable end-to-end autonomous coding workflows.

Addressing the alignment and optimization of these systems remains a critical challenge. For alignment, methods like reward shaping and preference optimization are employed to mitigate bias and prevent catastrophic forgetting \cite{park2024spo}. Concurrently, a new paradigm of LLMs-as-optimizers has emerged. These models use evolutionary search and self-reflection to automatically refine objectives, hyperparameters, and action plans, showcasing a powerful new direction for autonomous system improvement \cite{yang2024largelanguagemodelsoptimizers, shinn2023reflexionlanguageagentsverbal, hemberg2024evolvingcodelargelanguage}.

\section{Methods}
\label{sec:method}

In this paper, we propose an end-to-end framework for the generation and evaluation of diffusion acceleration codes. We design two complementary components: \emph{DiffBench}, a comprehensive benchmark suite, and \emph{DiffAgent}, a unified LLM-based framework for code generation and optimization. Together, these components enable rigorous evaluation of generated code performance and fully automated generation of high-quality, optimized diffusion acceleration code.

\begin{figure*}[htbp]
    \centering
    \includegraphics[width=0.85\linewidth]{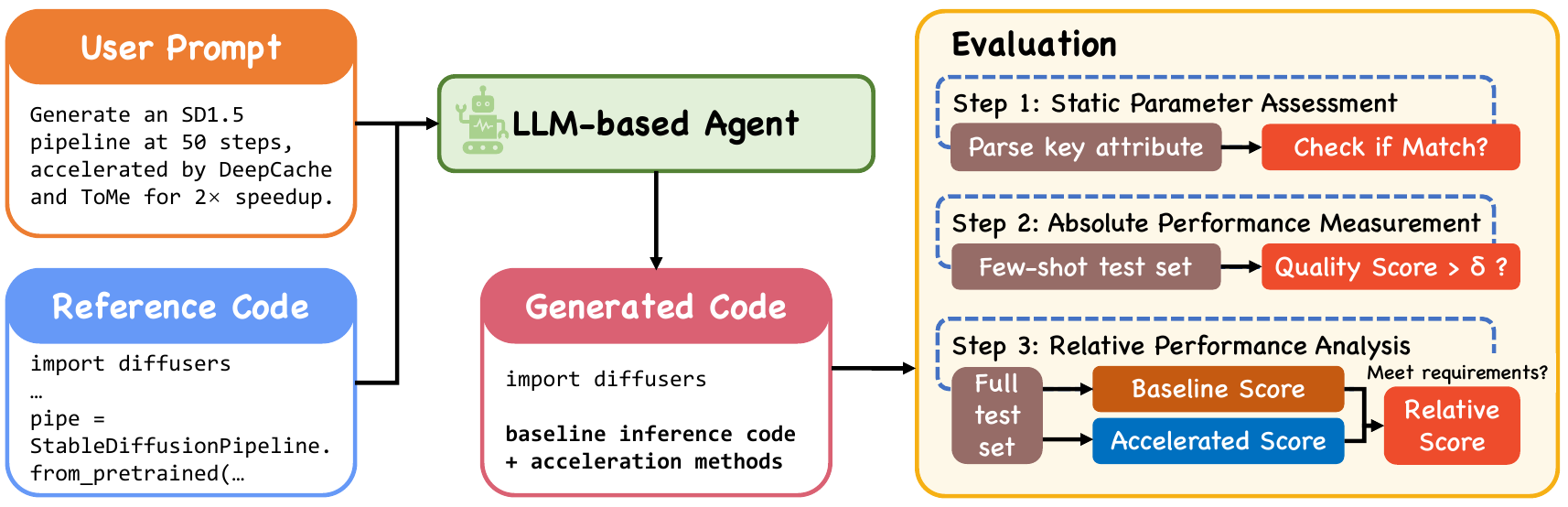}
    \caption{Overview of our proposed DiffBench. DiffBench tasks LLMs with generating diffusion acceleration code from a given user prompt and performs automated three-step evaluation.}
    \label{fig:diffbench}
\end{figure*}

\subsection{DiffBench: A Framework for Diffusion Acceleration Code Generation}
\label{sec:benchmark}
DiffBench is a new framework for evaluating the ability of language models to generate diffusion acceleration code that satisfies user requirements. In this section, we describe the task format, contents, and evaluation metric.

\paragraph{Task Format.}
DiffBench includes 604 tasks covering a range of diffusion acceleration scenarios and is easily extensible to new development contexts. The end-to-end specification of a task is shown in Figure~\ref{fig:diffbench} and described below.

\begin{itemize}
\item \textbf{Task Input:}  
The input to each task consists of a user prompt and reference code compiled from open source diffusion libraries. The user prompt may specify the diffusion model type, sampling steps, chosen acceleration methods, and the target speedup to be achieved on a specified hardware platform. The reference code implementations include the definition of the diffusion model instance, the code required to perform inference under specified conditions such as text or image inputs, and the associated preprocessing and postprocessing procedures.

\item \textbf{Task Output:}  
Given the task input, the language model must produce a new diffusion inference implementation that meets the user requirements. For example, the model may combine multiple acceleration techniques in order to achieve the required speedup while keeping the relative quality loss within an acceptable threshold.
\end{itemize}

To succeed, the model must determine (1) which acceleration methods in the diffusion sampling process will most effectively contribute to the target speedup and (2) how to tune the parameters of those methods. The model therefore requires a deep understanding of diffusion architectures and acceleration strategies, as well as experience in parameter tuning, in order to fulfill these objectives.

\paragraph{Task Definition.}
The 604 tasks in DiffBench are organized into five levels according to deployment complexity. Each level presents distinct challenges:

\begin{itemize}
  \item \textbf{Level 1 (41 tasks):} Basic pipeline generation. Tasks at this level require the model to construct a standard diffusion inference pipeline without any acceleration techniques. Defining the correct model type and architecture is challenging due to the diversity of existing diffusion frameworks.  

  \item \textbf{Level 2 (116 tasks):} Single‑method acceleration. These tasks require the addition of a single acceleration technique, chosen from a library of methods such as fast samplers, feature reuse or token merging. Although documentation for individual acceleration techniques is mature, correctly integrating a specified method into a particular pipeline remains difficult. An inappropriate choice can over‑accelerate the model and produce noisy outputs.  

  \item \textbf{Level 3 (261 tasks):} Compositional acceleration. This level introduces combinations of multiple acceleration methods, for example, pairing faster samplers with feature reuse, to achieve greater speedup potential. The LM model must reason about how to combine techniques effectively without compromising output quality.  

  \item \textbf{Level 4 (93 tasks):} Acceleration with explicit speedup target. Tasks at this level specify a minimum speedup requirement while constraining relative quality loss to within a threshold \(\delta\) (for example, \(\delta = 5\%\)). Models must iteratively adjust and debug their implementations to meet these quantitative criteria.  

  \item \textbf{Level 5 (93 tasks):} Acceleration with latency constraint. In addition to maintaining relative quality loss below \(\delta\), tasks at this level impose a maximum allowable inference latency on a given hardware platform. Both Level 4 and Level 5 demand repeated verification and debugging in the target environment to satisfy user requirements.  
\end{itemize}

\begin{figure}[htbp]
    \centering
    \begin{subfigure}[b]{0.48\linewidth}
        \centering
        \includegraphics[width=\linewidth]{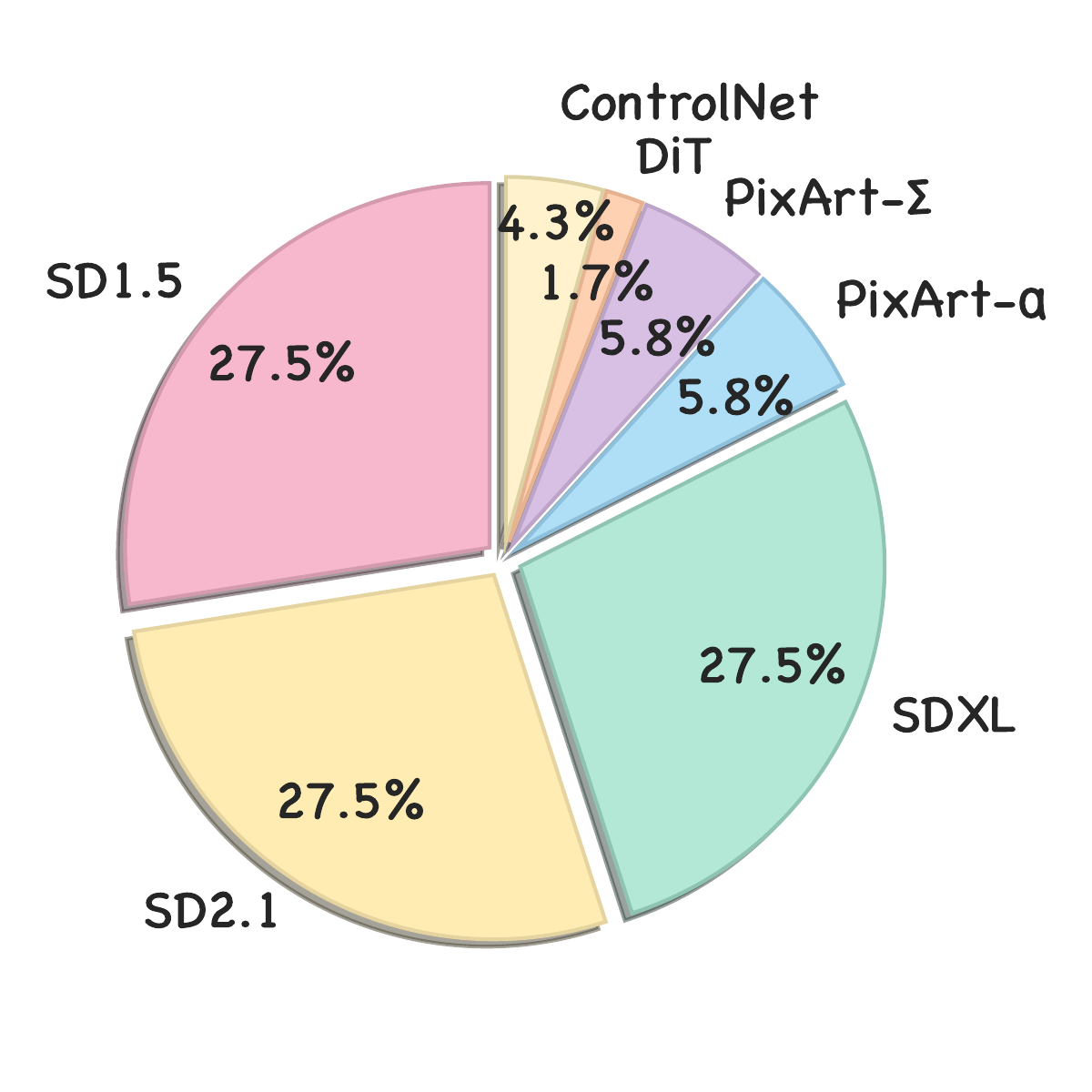}
        \label{fig:pipe_distribution}
    \end{subfigure}
    \hfill
    \begin{subfigure}[b]{0.48\linewidth}
        \centering
        \includegraphics[width=\linewidth]{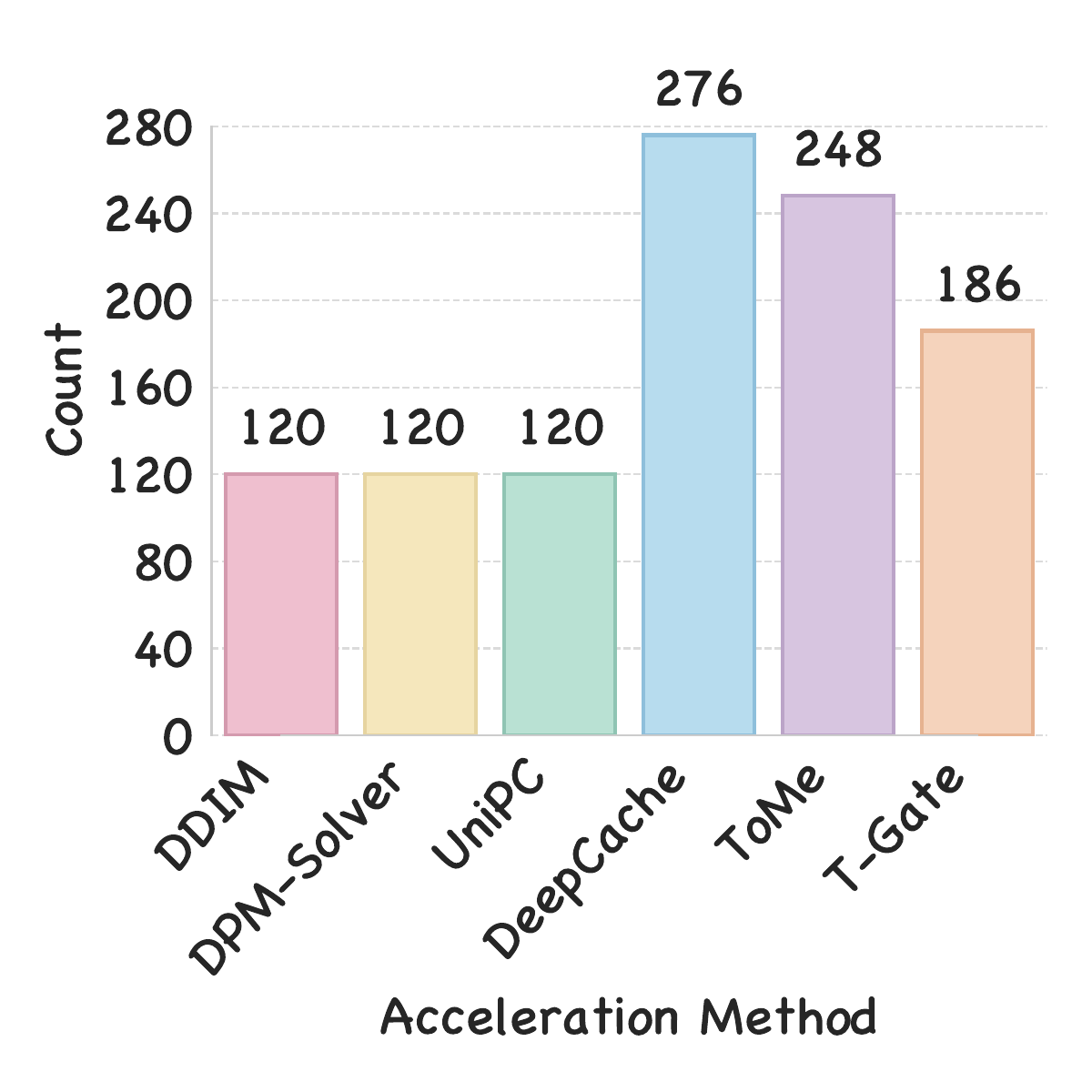}
        \label{fig:bar}
    \end{subfigure}
    \vspace{-15pt}
    \caption{Illustration of the pipeline and distribution of acceleration methods. Our benchmark covers diverse pipelines and acceleration strategies to enable comprehensive evaluation of LLM-driven acceleration code generation.}
    \label{fig:benchmark_distribution}
\end{figure}

\paragraph{Task Construction.}
We assemble a benchmark of 604 prompts paired with corresponding ground‐truth code implementations to evaluate acceleration strategies across a broad spectrum of diffusion models. The benchmark covers U‑Net backbones (SD1.5 \cite{rombach2022high}, SD2.1 \cite{rombach2022high}, SDXL \cite{podell2023sdxl}) and transformer variants (DiT \cite{peebles2023scalable}, PixArt‑\(\alpha\) \cite{chen2023pixartalpha}, PixArt‑\(\Sigma\) \cite{chen2024pixartsigma}). It supports text‑to‑image, class‑to‑image and image‑to‑image conditioning \cite{meng2021sdedit} and spans resolutions from \(256\times256\) to \(1024\times1024\). Three popular samplers (DDIM \cite{song2020denoising}, DPM‑Solver \cite{lu2022dpm}, UniPC \cite{zhao2023unipc}) are included along with four acceleration techniques: token merging (ToMe \cite{bolya2023token}), feature reuse (DeepCache \cite{ma2024deepcache}), gated activation (T‑Gate \cite{tgate}) and half‑precision computation (FP16). We present the distribution of pipeline types and explicitly required acceleration methods in the benchmark prompts. The results are shown in Figure \ref{fig:benchmark_distribution}.

During benchmark construction, for Levels 1--3 we generate baseline pipeline code and verify correctness and output quality. For Levels 4 and 5, we define an acceleration search space for each baseline and perform a 50-iteration search on a 36-sample validation set, aiming to maximize speedup under a CLIP-Score degradation bound \(\sigma\). Configurations meeting this bound are labeled as ``medium'' samples. We then scale the maximum speedup by factors \(\Delta_{1}\) and \(\Delta_{2}\) to generate ``easy'' and ``hard'' samples, noting that hard samples may not always have valid solutions.

\paragraph{Evaluation.}
To determine whether the generated code fulfills its intended purpose, we employ a three‑stage automated evaluation pipeline comprising static parameter assessment, absolute performance measurement and relative performance analysis. These stages increase in rigor and are tailored to different prompt categories in the benchmark.

\begin{itemize}
  \item \textbf{Static Parameter Assessment.}  
  In the first stage, we verify that all critical parameters in the candidate code match those in the ground-truth code while permitting additional non‑essential fields. We extract key attributes such as the pipeline class (e.g., \verb|StableDiffusionPipeline|), model identifier (e.g., \verb|stable-diffusion-v1-5|), scheduler class (e.g., \verb|DDIMScheduler|), number of inference steps, applied acceleration methods, target resolution and the use of specific preprocessors (for example, Canny edge detection). An automated parser extracts these fields from both the ground truth and the generated code, enforcing exact matches for ground‑truth components and tolerating extraneous code. Samples that pass this static check proceed to the next stage.

  \item \textbf{Absolute Performance Measurement.}  
   In the second stage, we evaluate the generated diffusion sampling code on a few‐shot set of 10 examples using the CLIP‐Score as a quality metric. Any sample scoring below a predefined threshold $\delta$ is marked as failing this evaluation. Evaluation samples are randomly selected from the COCO dataset \cite{lin2014microsoft}.

  \item \textbf{Relative Performance Analysis.}
    The third stage conducts a quantitative analysis of the generated code’s performance relative to a reconstructed baseline implementation, and it is invoked when specific metrics are required, for example inference speedup in Level 4 or raw latency in Level 5. For Level 4 tasks, we define two metrics, \(L\) and \(U\), as follows:
    \begin{equation}
    \label{eq:L}
    L = \frac{\frac{1}{N}\sum_{i=1}^{N}\bigl(S_{\mathrm{base}}^{(i)} - S_{\mathrm{acc}}^{(i)}\bigr)}
             {\frac{1}{N}\sum_{i=1}^{N} S_{\mathrm{base}}^{(i)}},
    \end{equation}
    \begin{equation}
    \label{eq:U}
    U = \frac{\frac{1}{N}\sum_{i=1}^{N} T_{\mathrm{base}}^{(i)}}
             {\frac{1}{N}\sum_{i=1}^{N} T_{\mathrm{acc}}^{(i)}},
    \end{equation}
    where \(N\) denotes the number of validation samples. For each sample \(i\), \(S_{\mathrm{base}}^{(i)}\) and \(S_{\mathrm{acc}}^{(i)}\) are the quality scores, i.e., CLIP scores, of the baseline and accelerated implementations, respectively. Likewise, \(T_{\mathrm{base}}^{(i)}\) and \(T_{\mathrm{acc}}^{(i)}\) denote the inference times under the baseline and accelerated implementations. A smaller value of \(L\) indicates stronger preservation of generative quality, while a larger value of \(U\) indicates greater inference efficiency. In Level 5 tasks, we substitute \(U\) with the raw latency measurement \(\tau\) obtained on the specified hardware platform.

\end{itemize}


Only samples that satisfy all three stages are considered to pass. Following previous works \cite{zhang2024codeagent, zheng2023codegeex}, we define the pass rate as the proportion of tasks for which the generated code is functionally correct and meets the user’s requirements. We focus primarily on \textbf{$S_p$}, which represents the probability that the top‑ranked submission is correct and reflects real‑world usage, where only one suggestion is typically adopted. For tasks labeled as “hard,” we introduce the achievement rate \(s_{a}\). Let $ \rho = \frac{U}{U_{\mathrm{req}}},$ where \(U\) is defined in Equation \eqref{eq:U} and \(U_{\mathrm{req}}\) denotes the required speedup. The \textbf{achievement rate} is then given by $ S_{a} = \min\{\rho,\,1\}.$ This formulation provides a continuous scoring mechanism for evaluating performance on demanding optimization tasks.

\subsection{DiffAgent: A Unified System for Code Generation and Optimization}
\begin{figure}[!htbp]
    \centering
    \includegraphics[width=1.0\linewidth]{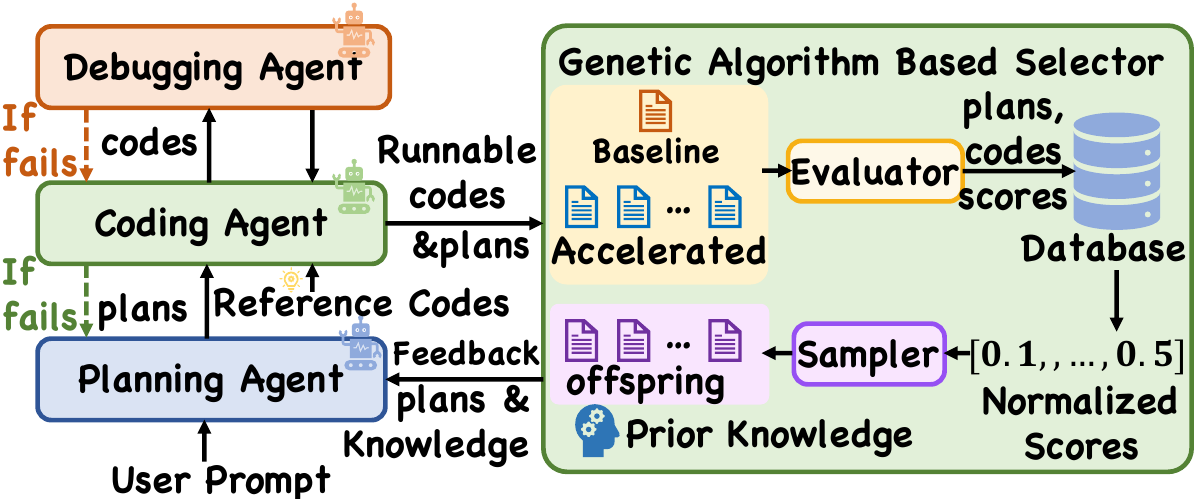}
    \caption{Overview of our proposed DiffAgent. Through multi-agent collaboration and genetic algorithm based optimization, DiffAgent generates high-quality diffusion acceleration code and iteratively refines it.}
    \label{fig:diffagent}
\end{figure}

We introduce DiffAgent, an LLM framework that unifies code generation and optimization to enhance the production of diffusion acceleration code. DiffAgent consists of four core component: planning agent, coding agent, debugging agent and genetic algorithm based selector. The overview pipeline is illustrated in Figure \ref{fig:diffagent}.

\paragraph{Planning Agent.}
The planning agent generates detailed code generation plans based on the user prompt.
For code generation tasks (Levels 1–3 in DiffBench), the planning agent formulates a coding plan and passes it to the coding agent directly. For optimization tasks (Levels 4 and 5 in DiffBench), the planning agent begins by creating a baseline plan that outlines the implementation of a reference version without any acceleration techniques. It then generates prompt-conditioned acceleration plans, which define the selection, ordering, and parameterization of optimization strategies needed to achieve the desired performance target.
As shown in Figure \ref{fig:diffagent}, during the optimization loop, the planning agent receives the generation plans of the $M$ promising offspring. For each of these, the generation plan is augmented with a feedback report and prior tuning insights (empirical knowledge for optimizing diffusion acceleration parameters). The planning agent uses this augmented context to produce up to $M$ refined, next-generation plans.  Concurrently, to maintain diversity and avoid converging to a local optimum, it generates $P - M$ entirely new plans based on the original user prompt. This produces a total of $P$ plans for the subsequent iteration. By default, $P = 7$ and $M = 4$. In scenarios where no optimization is required, both $P$ and $M$ are set to zero.


\paragraph{Coding Agent.}
The coding agent is used to generate diffusion inference code conditioned on each specified plan. Given the baseline code plan and $P$ acceleration code plans, it produces the baseline implementation and $P$ accelerated variants, respectively. To improve the model’s understanding of diffusion inference code structure, we compile templates of acceleration code as reference codes from established diffusion libraries. These reference codes are concatenated with each code plan before being provided to the coding agent. With this prior information, the agent can generate accurate and well‑structured diffusion acceleration code.

\paragraph{Debugging Agent.}
The debugging agent verifies the correctness of the generated code by leveraging the Reflexion \cite{shinn2023reflexion} architecture to identify errors and propose corrections. It coordinates with the coding agent to iteratively produce revised implementations for up to \(T_{\text{debug}}\)  iterations. If no runnable version is obtained after \(T_{\text{debug}}\)  attempts, the coding agent restarts code generation conditioned on the same plan (the {orange} arrow in Figure \ref{fig:diffagent}), and this regeneration process may be repeated for up to \(T_{\text{code}}\) restart cycles. If the code remains nonfunctional after these cycles, the framework backtracks to the planning stage (the green arrow in Figure \ref{fig:diffagent}), prompting the planning agent to generate a new plan. In the worst case, the system will invoke the LLM \(T_{\mathrm{code}} \times T_{\mathrm{debug}}\) times per user prompt. By default, we set \(T_{\text{code}} = 5\) and \(T_{\text{debug}} = 3\).

\paragraph{Genetic Algorithm Based Selector.}

After obtaining runnable code implementation, we employ a genetic algorithm-based selector to evaluate the performance of each implementation and guide the agent toward producing high-performance implementations. Specifically, we assess each implementation along two dimensions: quality and efficiency. We compute the relative quality loss as defined in Equation~\ref{eq:L} and the speedup as defined in Equation~\ref{eq:U}. These two metrics are combined using a weighted sum to produce a fitness score for each variant. Each code implementation, along with its corresponding generation plan and computed fitness score, is recorded in a database. If any implementation satisfies the user’s specified quality and efficiency requirements, the process terminates and the corresponding code is returned. Otherwise, we normalize the fitness scores to derive sampling probabilities and select $M$ implementations as promising offspring.

For each of these $M$ offspring, we use its associated generation plan to guide the next iteration. A feedback report is created to describe the current code's quality and efficiency, explicitly comparing it against the performance targets. This feedback, along with the original generation plan and prior tuning insights, is passed back to the planning agent to generate the next round of plans. The genetic optimization process is repeated for up to \(T_{\text{sel}}\) iterations, where \(T_{\text{sel}}\) is set to 5 by default, until the user’s requirements are satisfied.

\section{Experiments}

\begin{table*}[htbp]
  \centering
  \begin{tabular}{lcccccc}
    \toprule
    \textbf{Method} & \textbf{Level 1} & \textbf{Level 2} & \textbf{Level 3} & \textbf{Level 4} & \textbf{Level 5} & \textbf{Avg.} \\
    \midrule
    o3-mini                              & 41.46 & 24.14 &  4.60 &  9.68 &  6.45 & 11.92 \\
    Claude Sonnet 4                      & 78.04 & 72.41 & 76.25 &  5.38 &  8.60 & 54.30 \\
    GPT-4.1                              & 56.10 & 18.97 &  7.28 & 10.75 & 12.90 & 14.24 \\  
    Gemini 2.5 Flash                     & 39.02 & 29.31 &  7.66 &  2.15 &  1.08 & 12.09\\
    \rowcolor{gray!20}
    o3-mini w/ DiffAgent                 & 73.17$_{\mathbf{+31.71}}$ & 70.69$_{\mathbf{+46.55}}$ & 69.73$_{\mathbf{+65.13}}$ & 22.58$_{\mathbf{+12.90}}$ & 27.96$_{\mathbf{+21.51}}$ & 56.46$_{\mathbf{+44.54}}$ \\
    \rowcolor{gray!20}
    Claude Sonnet 4 w/ DiffAgent         & 90.24$_{\mathbf{+12.20}}$ & 91.38$_{\mathbf{+18.97}}$ & 99.23$_{\mathbf{+22.98}}$ & 33.33$_{\mathbf{+27.95}}$ & 63.44$_{\mathbf{+54.84}}$ & 81.59$_{\mathbf{+27.29}}$ \\
    \bottomrule
  \end{tabular}
  \caption{Comparison of pass rate $S_{p}$ on the DiffBench benchmark across various LLMs, reported per difficulty level. All values are percentages, gains relative to the corresponding baseline are shown as subscripts in \textbf{bold}.}
  \label{tab:baseline_comparison}
\end{table*}

We conducted extensive experiments to address three research questions: (1) How well do existing code‑generation LLMs perform when evaluated on DiffBench, and what challenges do they face? (2) To what extent does our DiffAgent outperform these baseline LLMs in generating diffusion‑model acceleration code? (3) How effectively do the individual modules we’ve integrated into the agent system support the code‑writing process?

\subsection{Experimental Setup}
\paragraph{Benchmark.} To evaluate our method on diffusion acceleration code generation, we conducted experiments on the DiffBench benchmark. We set \(\sigma = 5\%\), \(\Delta_{1} = 0.8\) and \(\Delta_{2} = 1.2\).

\paragraph{Base LLMs.} We evaluated the performance of several recent large language models on our benchmark: GPT-4.1, Claude Sonnet 4, Gemini 2.5 Flash, and o3-mini. These represent some of the most advanced models available at the time of our evaluation.

\subsection{DiffBench Baseline Evaluation}

In this section, we evaluate a variety of language models on DiffBench without further fine‑tuning to assess their out‑of‑the‑box performance and identify failure modes.

\paragraph{Baseline.}
We applied our custom diffusion‑model acceleration benchmark, DiffBench, to evaluate four leading LLMs, including both foundation models and inference-optimized models. DiffBench proves more challenging than existing suites: as shown in Table \ref{tab:baseline_comparison}, despite their strong performance on standard coding tasks, these models achieve pass rates below 35\% on our acceleration challenges.

\begin{figure}[h]
    \centering
    \begin{subfigure}[b]{0.48\linewidth}
        \centering
        \includegraphics[width=\linewidth]{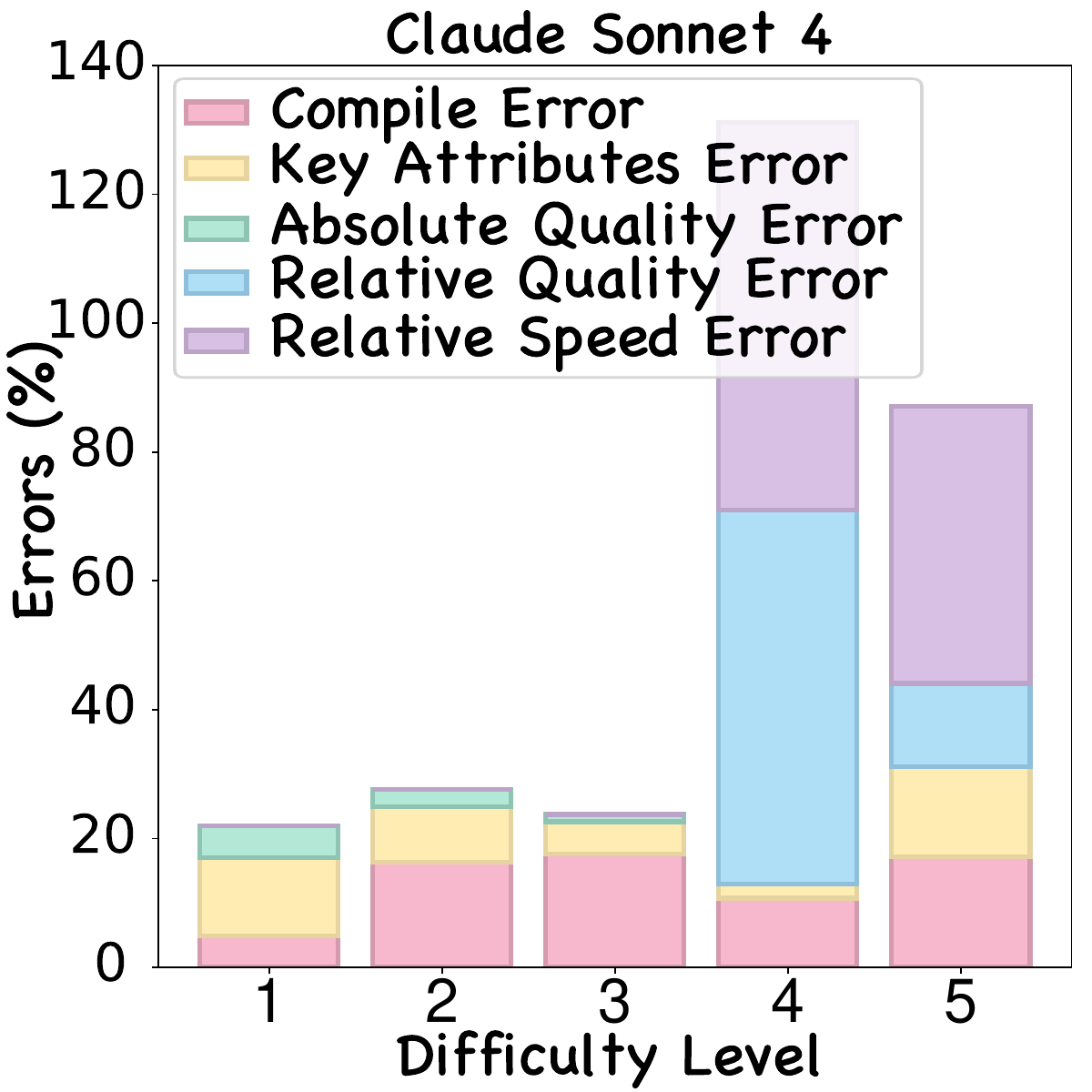}
        \label{fig:error_analysis}
    \end{subfigure}
    \hfill
    \begin{subfigure}[b]{0.48\linewidth}
        \centering
        \includegraphics[width=\linewidth]{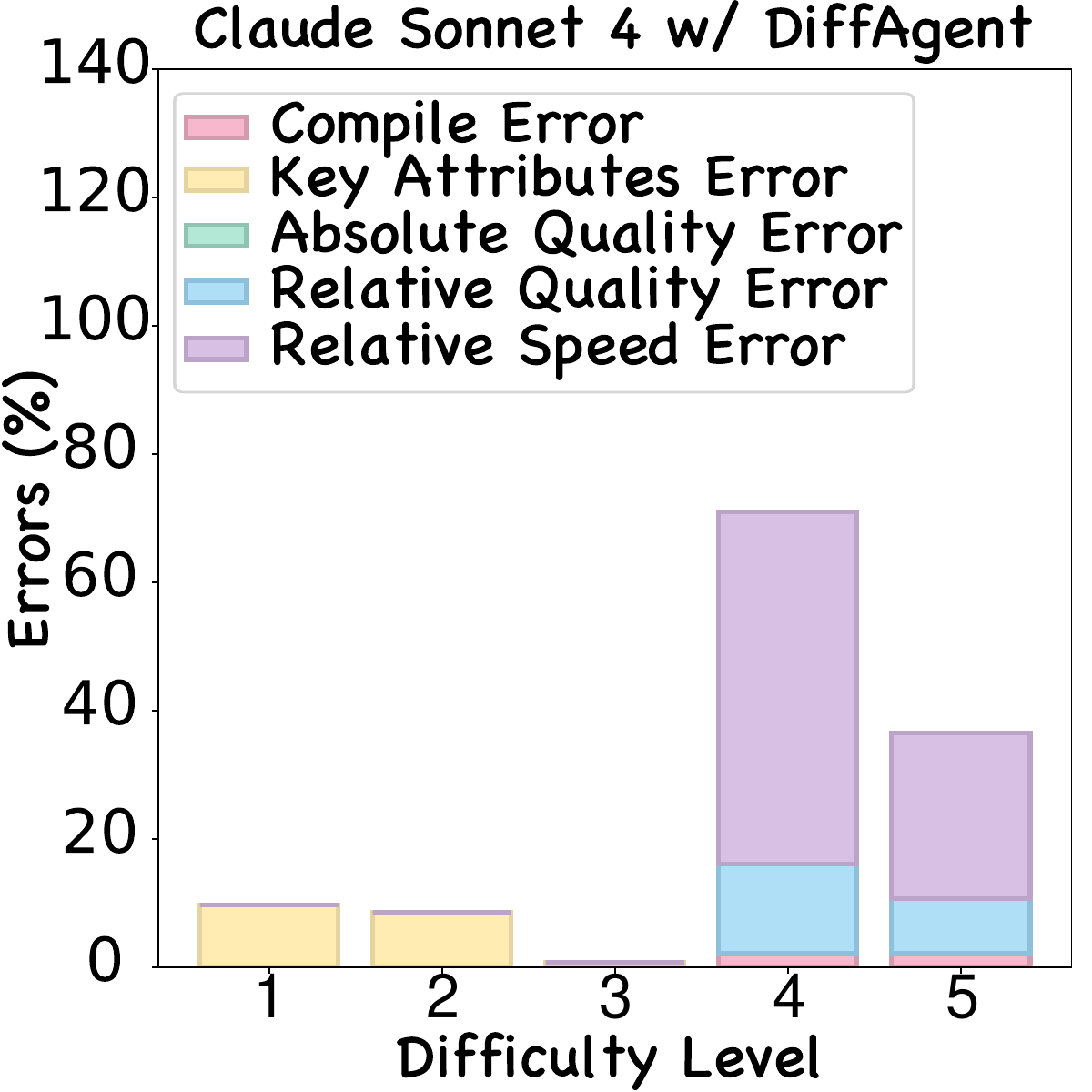}
        \label{fig:error_analysis_ours}
    \end{subfigure}
    \vspace{-15pt}
    \caption{Error mode analysis on the DiffBench benchmark across various LLMs, reported per difficulty level. We categorize failure modes of generated code into five failure models. All values are percentages.}
    \label{fig:error_comparison}
\end{figure}

\paragraph{Error Analysis.}

Figure~\ref{fig:error_comparison} provides a comprehensive breakdown of the failure modes in diffusion acceleration code generated by Claude Sonnet 4, both with and without DiffAgent integration. We classify these failures into five categories: Compile Error and Key Attributes Error, corresponding to evaluation stage 1; Absolute Quality Error for stage 2; and Relative Quality Error and Relative Speed Error for stage 3. Note that Relative Quality Error and Relative Speed Error are not mutually exclusive, so the sum of error rates may exceed 100\%.

Our results indicate that DiffAgent yields substantial improvements in addressing fundamental code generation issues. Across all difficulty levels, the rate of Compile Error is markedly reduced. For example, at Level 5, the Compile Error rate decreases from 31.18\% for the baseline Claude Sonnet 4 model to 2.15\% when DiffAgent is employed. In addition, the incidence of Key Attributes Error is significantly lower, and Low Quality Error is completely eliminated in the scenarios tested, dropping to 0\% in several cases.

When examining tasks that involve complex performance constraints (Levels 4 and 5), DiffAgent demonstrates a pronounced ability to balance competing objectives effectively. While the baseline model exhibits high rates of Relative Quality Error and Relative Speed Error, DiffAgent achieves a dramatic reduction in both error types at these higher difficulty levels. This finding underscores the capacity of DiffAgent to optimize for quality and speed simultaneously without compromising one objective for the other.

\begin{table*}[htbp]
  \centering
  \begin{tabular}{lcccccc|ccc}
    \toprule
     & \multicolumn{6}{c}{$S_p$} & \multicolumn{3}{c}{$S_a$ (Hard-task)} \\
    \cmidrule(lr){2-7} \cmidrule(lr){8-10}
    \textbf{Method} &
      \textbf{Level 1} & \textbf{Level 2} & \textbf{Level 3} &
      \textbf{Level 4} & \textbf{Level 5} &
      \textbf{Avg.} & \textbf{Level 4} & \textbf{Level 5} & \textbf{Avg.}  \\
     
    \midrule
    DiffAgent & 90.24 & 91.38 & 99.23 & 33.33 & 63.44 & 81.59 & 56.61 & 79.92 & 68.27 \\
    w/o Knowledge Base  & 82.93 & 74.14 & 77.78 & 26.88 & 47.31 & 64.90 & 39.54 & 52.33 & 45.94 \\
    w/o GA             & 90.24 & 91.38 & 99.23 &  4.30 &  4.30 & 67.88 &  6.28 & 10.03 & 8.16\\
    w/o Debugging Agent   & 87.80 & 77.59 & 81.99 & 31.18 & 33.33 & 66.23 & 56.99 & 67.40 & 62.02 \\
    \bottomrule
  \end{tabular}
  \caption{Ablation study of key components in DiffAgent.  
    We report pass rate $S_p$ across five task levels and hard-task achievement rates $S_a$ on Level 4 and Level 5.}
  \label{tab:full_module_ablation}
  \vspace{-10pt}
\end{table*}

\subsection{DiffAgent Coding Performance}
In these experiments, we employed our diffusion acceleration code generation agent, DiffAgent, to augment the capabilities of leading LLMs. The results, summarized in Table \ref{tab:baseline_comparison}, show that DiffAgent consistently delivers substantial performance improvements across all base models and scales. For the Claude Sonnet 4 model, we observe a maximum gain of 27.29\%. Overall improvements range from 54.30\% to 81.59\%, validating the effectiveness of our proposed approach. These findings demonstrate that the integrated modules and domain knowledge within DiffAgent provide the necessary guidance for LLMs to generate accurate diffusion-model acceleration code and to successfully address the challenges of diffusion acceleration tasks.

For the Claude Sonnet 4 model, we illustrate the performance gains achieved by integrating DiffAgent across five levels of benchmark code, as shown in Figure 3. The most significant improvements appear in the third and fourth categories, because DiffAgent more accurately incorporates acceleration methods and better understands their parameter semantics, enabling correct usage that balances speed and accuracy to meet the prompt requirements.

\subsection{Ablation Study}

\paragraph{Effectiveness of Each Component.}
To assess the individual impact of DiffAgent’s three modules: the Knowledge Base, Genetic Algorithm and Debugging Agent, we conducted a leave-one-out ablation on the Claude Sonnet 4 backbone and evaluated performance on the full benchmark (Table~\ref{tab:full_module_ablation}). Removing any module degrades performance: the average pass rate $S_{p}$ decreases from $81.59\%$ to a range of $64.90\%\text{–}67.88\%$, and the hard-task achievement rate $S_{a}$ falls from $68.27\%$ to between $8.16\%$ and $62.02\%$.
(1) The Genetic Algorithm module is essential for complex tasks. Without it, pass rates on Levels 4 and 5, with $S_{p}$ dropping to $4.30\%$ and $S_{a}$ to $8.16\%$, this result underscores the role of GA in exploring and refining candidate programs for challenging problems.
(2) The Knowledge Base provides broad benefits across all levels. Its removal causes the largest absolute decline in overall pass rate, a $16.69\%$ reduction, demonstrating that domain knowledge accelerates search and improves solution quality. (3) The Debugging Agent primarily enhances robustness on high-difficulty tasks: without it, Level 5 pass rate declines by approximately $30\%$ and $S_{a}$ by $6.25\%$, showing that iterative error analysis prevents cascading mistakes. 
Collectively, these findings confirm that the three modules operate synergistically to deliver DiffAgent’s state-of-the-art performance.

\begin{table}[htbp]
  \centering
  \vspace{-5pt}
  \setlength{\tabcolsep}{6pt}
  \begin{tabular}{l|cc|cc|cc}
    \toprule
    & \multicolumn{2}{c|}{$T_\mathrm{sel}=2$} & \multicolumn{2}{c|}{$T_\mathrm{sel}=4$} & \multicolumn{2}{c}{$T_\mathrm{sel}=6$}\\
    \textbf{$P$} & $S_{p}$ & $S_{a}$ & $S_{p}$ & $S_{a}$ & $S_{p}$ & $S_{a}$ \\
    \midrule
    $4$  & 18.28 & 32.41 & 26.88 & 45.42 & 31.18 & 59.91 \\
    $7$  & 30.11 & 50.08 & 33.33 & 63.44 & 37.63 & 64.80 \\
    $10$ & 26.88 & 59.78 & 37.63 & 66.48 & 38.71 & 66.28 \\
    \bottomrule
  \end{tabular}
  \caption{Comparison of $P$ and $T_\mathrm{sel}$.}
  \label{tab:pop_iter_ablation}
  \vspace{-12pt}
\end{table}
\begin{figure}[!htbp]
    \centering
    \includegraphics[width=1.0\linewidth]{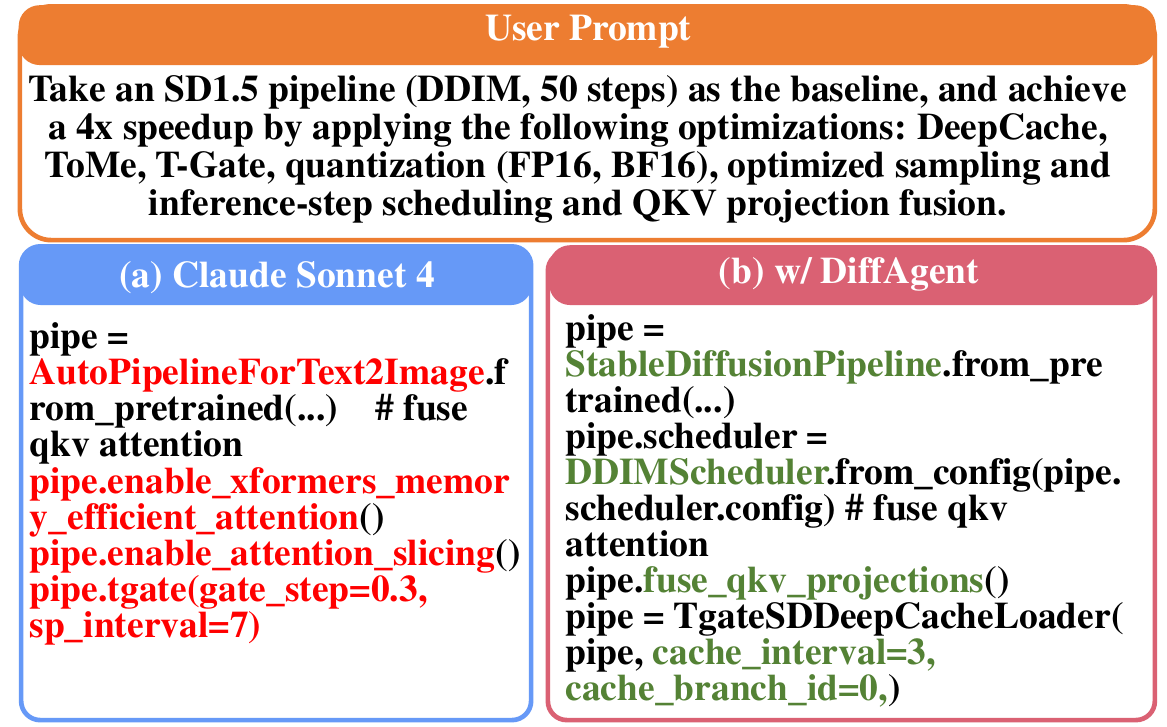}
    \caption{Sample visualization of code generation results. Incorrect code segments originally generated by the LLM are highlighted in red, while corrected segments produced by DiffAgent are shown in green.}
    \label{fig:vis}
\end{figure}

\paragraph{Comparison of GA Hyperparameters.}
To investigate the impact of two key genetic algorithm hyperparameters, population size $P$ and number of iterations $T_\mathrm{sel}$, on code generation, we conducted experiments on the fourth and fifth levels of DiffBench. The results in Table \ref{tab:pop_iter_ablation} indicate that performance plateaus once the population size exceeds 7 and the generation count exceeds 4. Beyond these thresholds, further increases yield negligible gains. To balance efficacy and runtime, we therefore set $P$ to 7 and $T_\mathrm{sel}$ to 4.

\subsection{Qualitative Analysis}
Figure \ref{fig:vis} contrasts code generated directly by the LLM with the version refined by DiffAgent. Our approach not only produces code that more faithfully meets the prompt, exhibiting consistent style and correct key attributes (e.g., pipeline, sampler), but also integrates acceleration techniques like DeepCache and T-Gate, making the resulting diffusion-acceleration code immediately applicable to real-world development scenarios.

\vspace{-10pt}
\section{Conclusion}
We introduce DiffBench, the first benchmark for LLM-generated diffusion pipelines, and DiffAgent, an LLM-driven automation framework for their acceleration. Evaluations show DiffAgent delivers substantial speedups with minimal quality loss across diverse scenarios, surpassing leading LLMs on complex and latency-sensitive tasks.

\bibliography{aaai2026}
\newpage

\newpage

\end{document}